%% file: main.tex
\documentclass[10pt,twocolumn,letterpaper]{article}

\usepackage{cvpr}              
\usepackage[accsupp]{axessibility} 

\usepackage{times}
\usepackage{epsfig}
\usepackage{graphicx}
\usepackage{amsmath}
\usepackage{amssymb}
\usepackage{pifont}
\usepackage{booktabs}
\usepackage{float}
\usepackage{adjustbox}
\usepackage{array}

\usepackage[pagebackref,breaklinks,colorlinks]{hyperref}
\usepackage{capt-of,etoolbox}

\usepackage[capitalize]{cleveref}
\crefname{section}{Sec.}{Secs.}
\Crefname{section}{Section}{Sections}
\Crefname{table}{Table}{Tables}
\crefname{table}{Tab.}{Tabs.}

\newcolumntype{C}[1]{>{\centering\let\newline\\arraybackslash\hspace{0pt}}m{#1}}

\newcommand{\fref}[1]{Fig.~\ref{#1}}
\newcommand{\tref}[1]{Table~\ref{#1}}

\newcommand{\aref}[1]{the supplementary materials}
\newcommand{\ETAL}[1]{~\etal~\cite{#1}}

\newcommand{\PLH}{{\mkern-2mu\times\mkern-2mu}}
\newcommand{\xtimes}[1]{${#1}\PLH{#1}$}

\newcommand{\email}[1]{\tt\small #1}
\newcommand{\inst}[1]{\textsuperscript{#1}}
\newcommand{\introfig}{\protect\centering\vspace{-5mm}
\includegraphics[width=\linewidth]{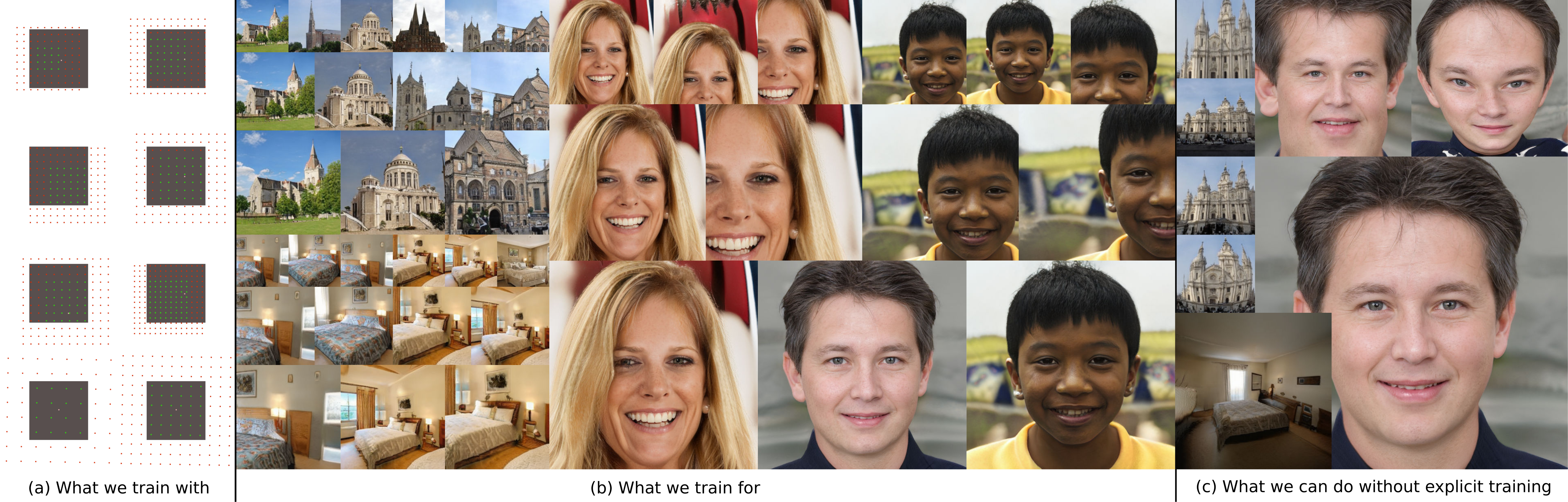}
\captionof{figure}{(a) We train with our scale-consistent positional encodings and modified generator architecture that enables (b) synthesis of arbitrary scales and multi-scale consistency. (c) We showcase our results for novel configurations not encountered during training such as generation at never-seen-before scales, extrapolation and editing using spatial transformations such warping and stretching.}%
\label{fig:intro}\vspace{6mm}}

\def\cvprPaperID{****} 
\def\confName{CVPR}
\def\confYear{2022}

\makeatletter
\apptocmd\@maketitle{{\introfig{}\par}}{}{}
\makeatother

\begin{document}

\title{Arbitrary-Scale Image Synthesis} 

\author{
Evangelos Ntavelis \inst{1,2}
\and
Mohamad Shahbazi\inst{1}
\and
Iason Kastanis\inst{2}
\and
Radu Timofte\inst{1}
\and
Martin Danelljan\inst{1}
\and
Luc Van Gool\inst{1,3}
\and
\\
\inst{1} Computer Vision Lab, ETH Zurich, CH 
\inst{2} Robotics \& ML, CSEM, CH 
\inst{3} KU Leuven, BE
\\
\email{entavelis,mshahbazi,radu.timofte,martin.danelljan,vangool@vision.ee.ethz.ch}\and
}
\maketitle

\begin{abstract}
Positional encodings have enabled recent works to train a single adversarial network that can generate images of different scales.
However, these approaches are either limited to a set of discrete scales or struggle to maintain good perceptual quality at the scales for which the model is not trained explicitly.
We propose the design of scale-consistent positional encodings invariant to our generator's layers transformations.
This enables the generation of arbitrary-scale images even at scales unseen during training. 
Moreover, we incorporate  novel inter-scale augmentations into our pipeline and partial generation training to facilitate the synthesis of consistent images at arbitrary scales. 
Lastly, we show competitive results for a continuum of scales on various commonly used datasets for image synthesis.
\end{abstract}\vspace{-3mm}


\input{source/introduction}
\input{source/relatedwork}

\input{source/method}
\input{source/experiments}

\input{source/conclusion}

\noindent\textbf{Acknowledgements}
This work was partly supported by CSEM and the ETH Future Computing Laboratory (EFCL), financed by a gift from Huawei Technologies.

{\small
\bibliographystyle{ieee_fullname}
\bibliography{eglib}
}

\end{document}


\title{Arbitrary-Scale Image Synthesis - Supplementary Material} 

\maketitle

\input{source/societal_impact}
\input{source/limitations}

\section{The generator's architecture}
\begin{figure}[t]
    \centering%
         \includegraphics[width=1.00\linewidth, trim=30 150 580 40, clip]{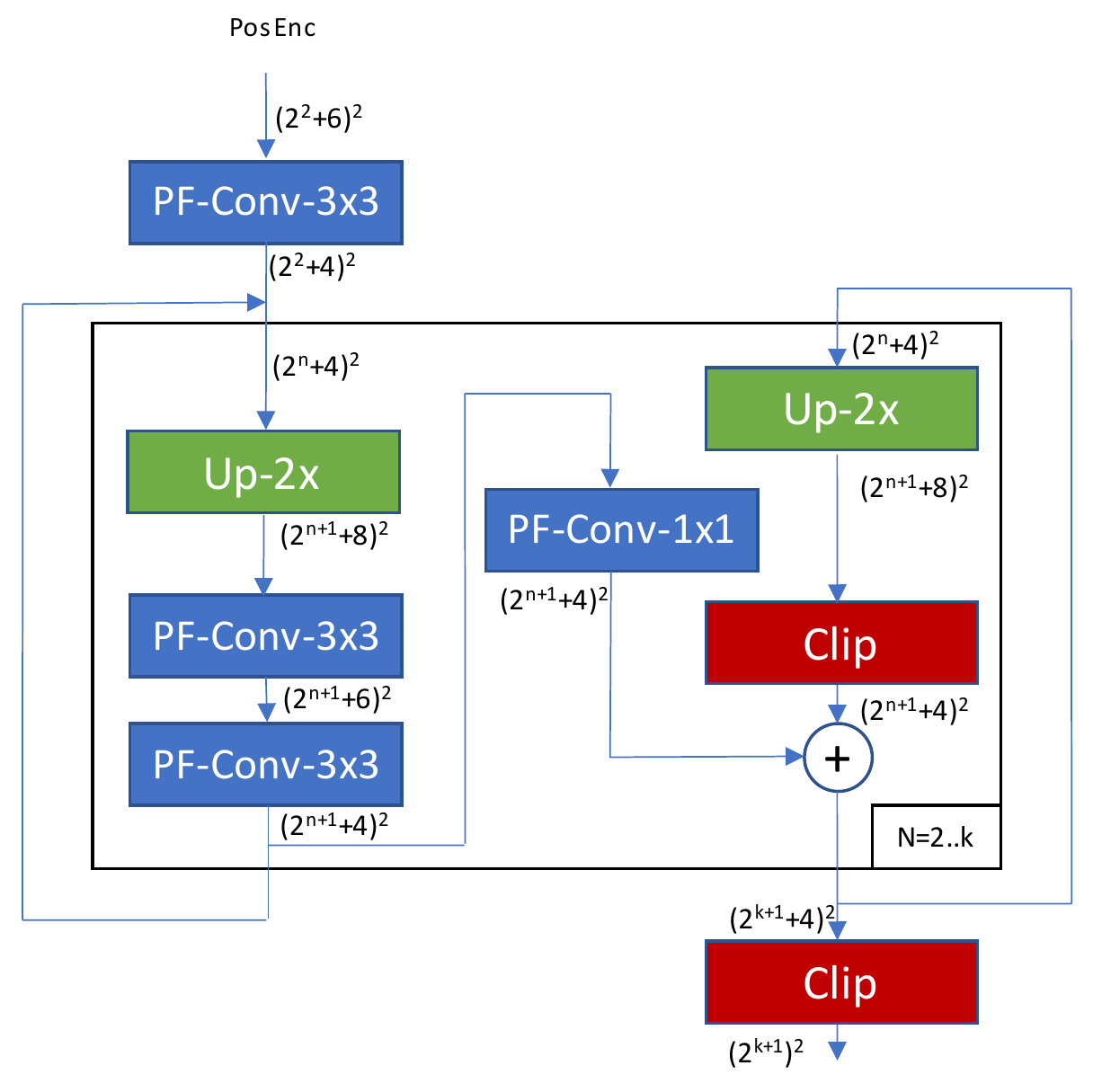} 
    \caption{Our generator's architecture. The blue blocks show the padding-free style modulated convolutions. The green blocks denote the operations of bilinear upsampling without aligning corner. The red blocks show the removal of the excess pixels introduced by the input padding in order for the feature maps and the output to match. }
    \label{fig:featmap}
 
\end{figure}

In \fref{fig:featmap} we can see a schematic of ScaleParty's generator. 

\section{Multi-scale training policies}

In this section we discuss the different scale training policies for FFHQ\cite{Karras_2019_CVPR} that we and the methods we compare with deploy:
\begin{itemize}
    \item CIPS\cite{anokhin2020cips} is trained with one target scale: 256.
    \item MS-PE\cite{Choi2021ICCV} is trained for 256, 320, 384, 448, 512.
    \item MSPIE\cite{mspie} is trained for 256, 384, 512. We use the version of MSPIE with cartesian spatial grid encodings, as it performs the best in terms of FID. Other encoding configurations exhibit similar behavior in the scales they were not trained for.
    \item We deploy the same setting as MSPIE for ScaleParty-noSC/Full and train for 256, 384 and 512.
    \item Our ScaleParty-Full which is trained with the scale consistency objective is trained with output resolutions of 256 and 384, but it can perform well even in higher resolutions.
    \item Our ScaleParty is also trained with output resolutions of 256 and 384. However, in contrast to all aforementioned approaches this is trained for partial generation; the generator is tasked to synthesized a multitude of scales. For example, during training it is generating 384 pixel parts of a 512 resolution full face picture. 
\end{itemize}

For LSUN Datasets\cite{yu15lsun}, we trained MSPIE, ScaleParty-noSC/Full and ScaleParty with outputs of 128 and 192. 

In order to facilitate faster and efficient training, we train our scale consistent versions of ScaleParty by continuing from an earlier checkpoint of the ScaleParty-noSC/Full version.

\section{Visual Results}

In this section we show a qualitative comparison between the state-of-the-art methods and different versions of ScaleParty.

\begin{figure*}[t]
    \centering%
     \includegraphics[width=1.\linewidth]{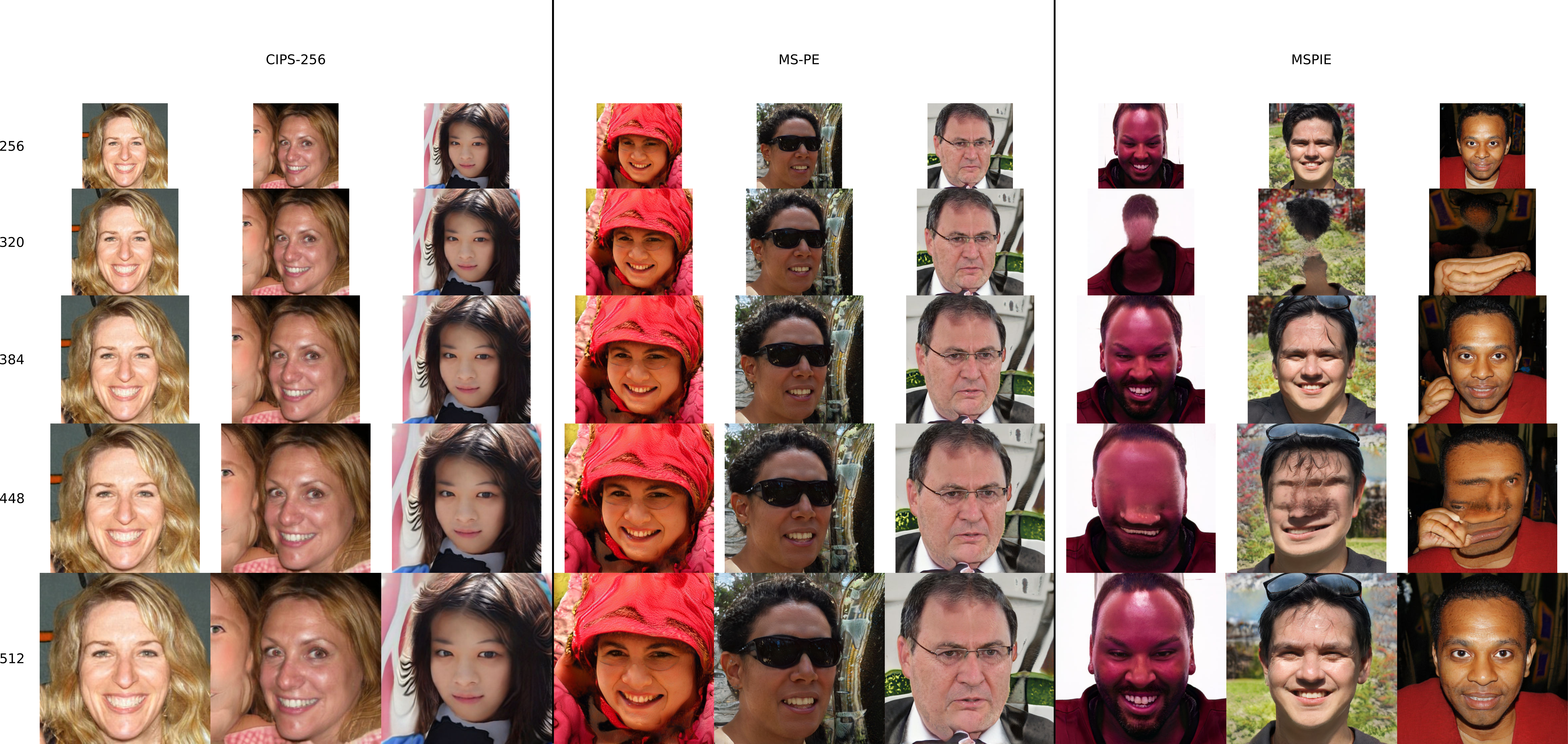} 
    \caption{Qualitative results of state-of-the-art methods on FFHQ\cite{Karras_2019_CVPR}. All images were picked randomly and generated without using the truncation trick. We find that the generated results from CIPS\cite{anokhin2020cips} and MS-PE\cite{Choi2021ICCV} exhibit a lot more artifacts than MS-PIE and our methods. However, note that MSPIE, while it performs the best in terms of FID among all methods, it is unable to generate in scales it was not trained for and it is the least consistent between the scales it generates.}
    \label{fig:sota}
\end{figure*}

\begin{figure*}[t]
    \centering%
     \includegraphics[width=1.\linewidth]{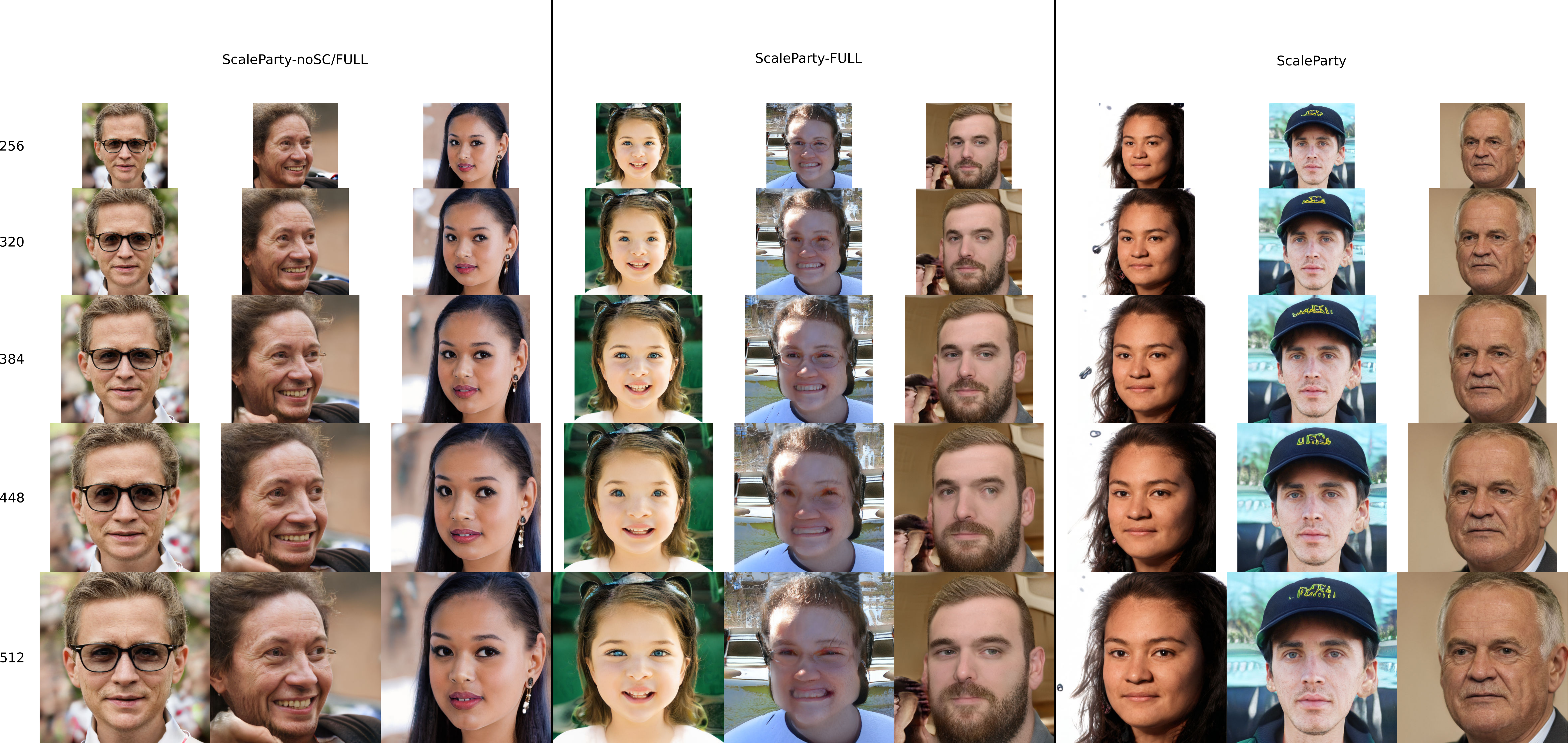} 
    \caption{Qualitative results of different versions of our method on FFHQ\cite{Karras_2019_CVPR}. All images were picked randomly and generated without using the truncation trick. We note that by drawing more samples the amount of generated images by our models that exhibit visual artifacts is comparable with MSPIE\cite{mspie} for the scales it was trained for, as it is also supported in the FID calculation. For our methods, we observe that only ScaleParty is able to generate results that are consistent, even for arbitrary scales.}

    \label{fig:ours}
\end{figure*}

\textbf{FFHQ\cite{Karras_2019_CVPR}:} In \fref{fig:sota} we can see visual results of the pretrained models of CIPS\cite{anokhin2020cips}, MS-PE\cite{Choi2021ICCV} and \cite{mspie}. In \fref{fig:ours} we can see the results for ScaleParty-noSC/Full, ScaleParty-Full and ScaleParty. While FID is lower for most scales for the versions trained with only full images, we can observe that the network applies a peculiar effect on the eyes of the faces it generates in scales it did not train for. We can see that both applying the scale consistency objective and partial generation is important for achieving consistent synthesis in arbitrary scales.

Moreover, in \fref{fig:arbitrary} we can see images synthesized at arbitrary scales. As the generator can only output certain resolutions, for scales between them, we generate at a higher resolution and crop the relevant part.

\begin{figure*}[t]
    \centering%
     \includegraphics[width=1.\linewidth]{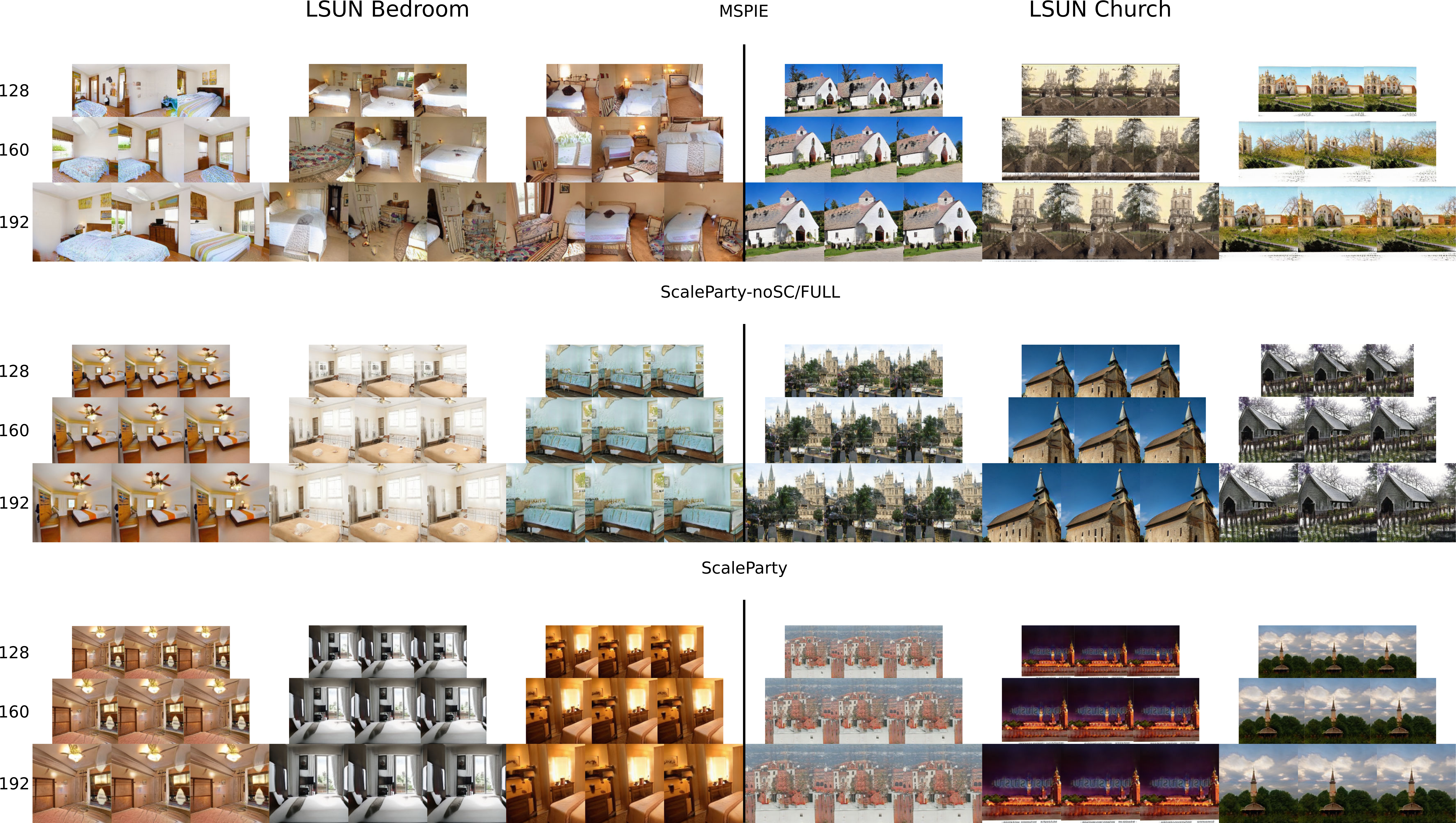} 
    \caption{Qualitative results on LSUN Bedroom and Church datasets. All images were picked randomly and generated without using the truncation trick.}
    \label{fig:lsun}
\end{figure*}

\textbf{LSUN~\cite{yu15lsun}:} In \fref{fig:lsun} we can see the qualitative results of MSPIE\cite{mspie}, ScaleParty-noSC/Full and ScaleParty trained for LSUN Bedroom and Church datasets. Note, that in combination with the weaker positional prior that these datasets have compared to FFHQ, we further augment this disparity by applying random cropping as a preprocessing step. Compared with FFHQ, MSPIE is generating more coherent results in the intermediate scale. However, in the case of LSUN Bedroom we can observe that the results are not consistent among different scales. 

In \fref{fig:lsun}, we visualize multiple syntheses of the same latent code and scale but with resampling the injected noise. We observe that ScaleParty is the most consistent among runs, while for MSPIE trained for LSUN Bedroom, we see that noise affects the generated images structurally.  

\begin{figure*}[t]
    \centering%
     \includegraphics[width=1.\linewidth]{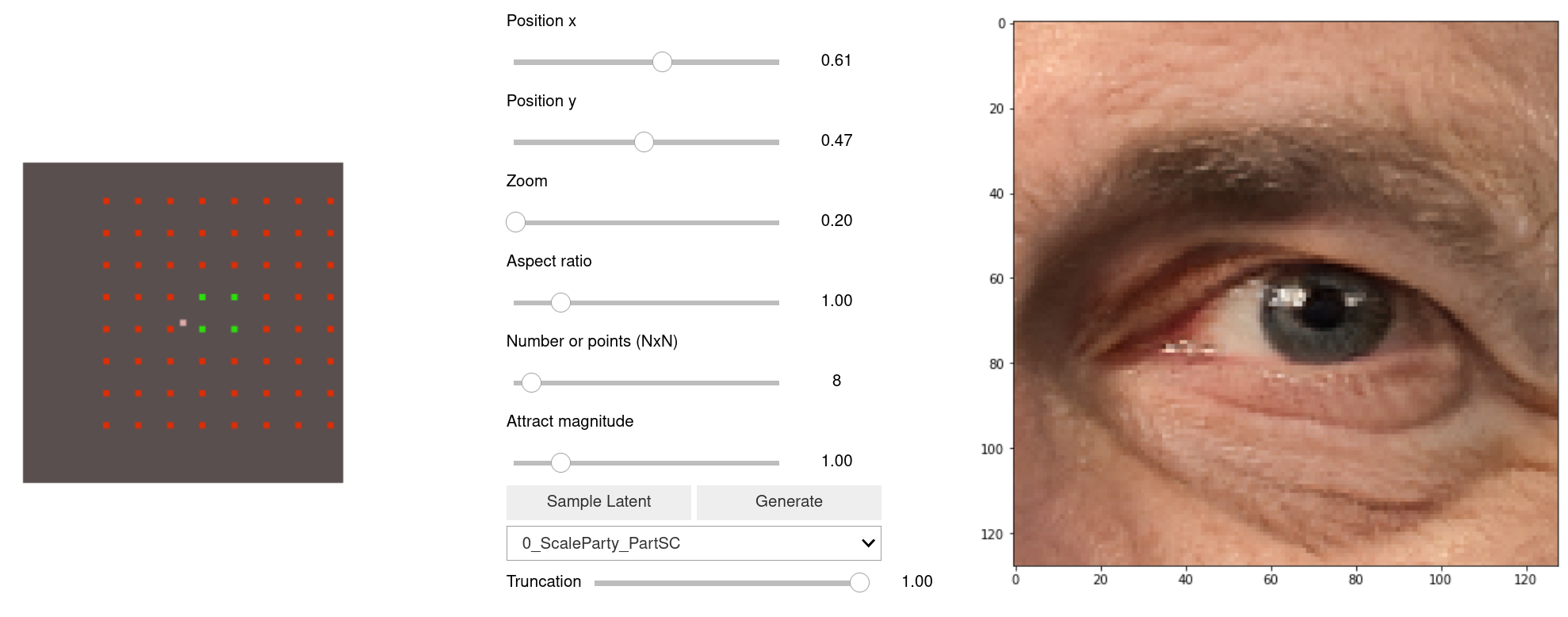} 
    \caption{The interface we developed to geometrically manipulate the generation by applying transformations to the positional encodings.}
    \label{fig:gui}
\end{figure*}

\begin{figure*}[t]
    \centering%
     \includegraphics[width=1.\linewidth]{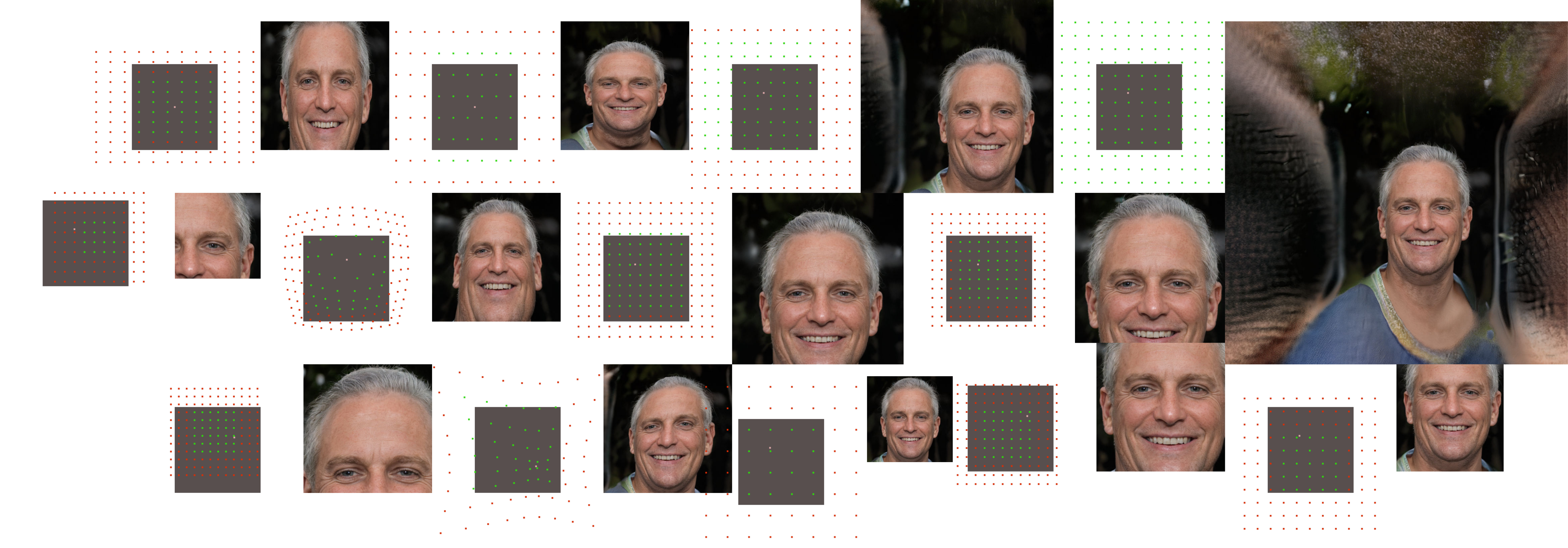} 
    \caption{Using our tool, we generate various images using the same latent code. The generated images are connected on their upper left corner with the positional encodings used to guide them. Changing the layout of the input yields different scales, resolutions and transformations. The gray box indicates the area the full face should occupy. The green dots show the actual area of the image space that is generated, while the red ones indicate the positional padding of the input, we utilize to counter the shrinking effect of padding-free convolutions.}
    \label{fig:manipulations}
\end{figure*}

\section{UI tool for guided generation}

We developed an interactive graphical user interface that permits the user to change the location, zoom factor, size of input, aspect ratio and warping of the positional encodings to guide the generation. Please refer to the accompanied video for more details. The tool will be available along with the code and the pretrained models.

In \fref{fig:manipulations} we showcase how changing the position, number and layout of the positional encodings can affect the generated image.

{\small
\bibliographystyle{ieee_fullname}
\bibliography{eglib}
}

%% file: source/introduction.tex
\section{Introduction}
\label{sec:introduction}



Generative adversarial networks (GANs)~\cite{goodfellowGAN} are the most commonly used paradigm for generating and manipulating images and videos~\cite{romero2019smit, viazovetskyi2020stylegan2,ververas2019slidergan, ntavelis2020sesame,hong2018learning,park2019SPADE,oasis}. The promising results obtained by GANs have motivated several applications of computer graphics and visual content generation. Ideally, a GAN model is not only capable of generating images similar to the training data but also provides the flexibility to manipulate and control the generation process for the target application~\cite{patashnik2021styleclip, jahanian2020steerability}. For instance, a GAN model used for animations and videos should be able to generate objects in different positions, scales, and viewpoints while maintaining consistency over other attributes of the object. Having a single model that provides control over different object attributes has received substantial attention from the research community~\cite{Collins2020EditingIS,explaining_in_style,li2021dystyle}. However, most of the existing GAN models are limited to the positional priors of their training data, making them unable to generate unseen translations and scales.

Xu~\etal~\cite{mspie} recently revealed that convolutional GANs learn the positional priors of their training data by using the zero paddings in the convolutions as an imperfect and implicit positional encoding. Motivated by such discovery, explicit positional encodings have been proposed to make the GAN models equivariant to different translations, scales, and resolutions~\cite{mspie, Choi2021ICCV, anokhin2020cips, inr_gan}. Positional encodings have created the possibility of obtaining a single GAN model, that can generate images with different resolutions, as well as different object scales and positions. However, despite the new opportunities brought about by the recent works, the existing methods are still limited to multi-scale generation only in discrete resolutions. They suffer from object inconsistency between different scales and resolutions.

To address the aforementioned limitations, we aim to extend the task of multi-scale generation, using a single generator, to \emph{arbitrary continuous} scales. To this end, we first propose a more suitable positional encoding formulation. While this leads to arbitrary-scale generation, this strategy alone does not guarantee consistency across scales. We therefore further propose a means of enforcing consistency between different scales and resolutions using inter-scale augmentations in the discriminator. Specifically, we generate images at different scales from the same latent code. Then, pairs of generated images at different scales go through channel-mix and cut-mix augmentations. Finally, the discriminator classifies the augmented images as real or fake. Such an approach encourages the generator to generate scale-consistent images so that the images still look realistic after inter-scale augmentations.
Lastly, our method can also generate parts of the image in arbitrary resolutions with scale consistency, as visualized in Figure~\ref{fig:intro}. 

To summarize our contributions:
\begin{itemize}
    \item We design a scale-consistent positional encoding scheme that enables fully convolutional and pad-free generators to generate images of arbitrary scales. 
    \item We introduce a set of inter-scale augmentations that pushes the generator to create consistent images among scales.
    \item We further facilitate the consistency among arbitrary scales  by incorporating partial generation in our training pipeline.
\end{itemize}
We perform experiments on various commonly used datasets characterized by diverse positional priors. 
Our results indicate that the introduced pipeline permits the consistent generation of images of arbitrary scales while preserving high visual quality.

%% file: source/relatedwork.tex
\section{Related Work}
\label{sec:related}

Generative adversarial networks have been exploited in various applications for unconditional generation~\cite{Karras_2019_CVPR}, as well as generation constrained by conditions, such as images~\cite{CycleGAN2017}, semantic categories~\cite{brock2018large, shahbazi2021cGANTransfer, shahbazi2022collapse}, semantic layouts~\cite{park2019SPADE, ntavelis2020sesame}, and text~\cite{pmlr-v48-reed16}. As the main focus of this work, existing methods on partial generation and multi-scale generation based on GANs are discussed in this section.

\subsection{Partial Generation} 
Standard GANs are usually trained to directly map a latent code to a full image. Models capable of partial generation, on the other hand, typically generate different parts of the image independently, which they can then aggregate to construct the full image. As investigated in previous works, partial generation can be posed as both patch-wise~\cite{cocoganLin_2019, cheng2021inout, lin2021infinity, skorokhodov2021aligning, struski2020locogan} or pixel-wise~\cite{anokhin2020cips} generation of the images. The main  challenge in partial generation is maintaining the global structure and consistency of the full image. Therefore, position-aware generation using implicit or explicit positional encoding has become a crucial component of partial generation. Positional encodings have also been used in the context of semantic image synthesis\cite{tan2021efficient}.

COCO-GAN~\cite{cocoganLin_2019} generates different patches of the image and concatenates them to form the full image. The global consistency is ensured by using a generator that uses positional encodings coupled with a global latent code and a discriminator that assesses the quality of the concatenated patches. Infinity-GAN~\cite{lin2021infinity} is another model based on patch generation that combines a local latent code with global latent code and the positional encoding to drive generation. ALIS~\cite{skorokhodov2021aligning} exploits patch generation to generate images infinitely extendable in the horizontal direction.

INR-GAN~\cite{inr_gan} and CIPS~\cite{anokhin2020cips} differ from the aforementioned works as they perform partial generation pixel-wise. Instead of generating image patches using a convolutional network, they exploit fully-connected implicit neural representation (INR) to generate each pixel based on their position in the coordinate grid.
The sample-specific parameters of the INR for each image are generated by a hyper-network that receives the latent code as its input.

Contrary to these works, our generator learns global consistency  by generating smaller resolution full-frame images and imposing a multi-scale consistency objective.
 
\subsection{Multi-scale Generation}
Multi-scale generation can be defined as the task of generating images in different scales using a single model. MSG-GAN~\cite{lee2019maskgan} can be seen as one of the earlier works on multi-scale generation. Inspired by ProGAN~\cite{karras2018progressive}, the authors propose an architecture that outputs an RGB image at each layer of the generator, resulting in generating multiple scales of the same image. This approach, however, is only limited to the discrete resolutions up to the resolution of the final output. A recent study called MS-PIE~\cite{mspie}, proposes a padding-free fully-convolutional architecture capable of multi-scale generation based on the input positional encoding and the global latent code. The multi-scale generation can be done by feeding different resolutions of the positional encoding to the generator. To avoid shrinkage in the size of the padding-free feature maps, authors use bi-linear upsampling layers that generate feature maps with extra boundaries, compensating the lack of positional encoding. A similar recent study~\cite{Choi2021ICCV} achieves multi-scale generation by feeding the positional encodings at each layer of the generator, while retain the zero padding. We show that with proper design of positional encodings using them only as input is sufficient for multi-scale generation. Moreover, none of the aforementioned methods tackles the problem of synthesis at arbitrary scales nor addresses whether the multi-scale output is consistent.

CIPS~\cite{anokhin2020cips} and INR-GAN~\cite{inr_gan}, while trained for a single scale, are able to generate in multiple scale. Note, however, that their single-location conditional input does not contain any information about the scale they aim to generate in.

%% file: source/method.tex
\section{Our Method}
\label{sec:method}

\begin{figure*}[t]
    \centering
    \includegraphics[width=1.000\linewidth]{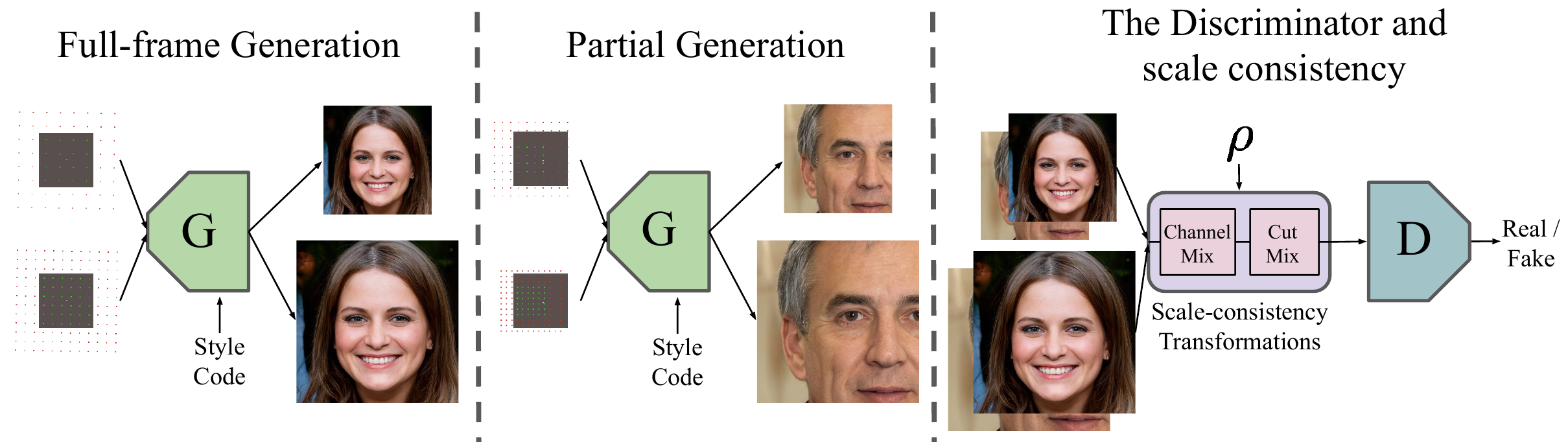}  
    \caption{Our training pipeline. We use positional encodings to guide the generation. Increasing their number leads to a larger \textit{resolution} while changing the spacing between them alters the \textit{scale}. The gray box indicates the positions corresponding to a full face. The number of red dots is constant and defines the positional padding used to compensate for the lack of zero padding in the generator. Applying interscale augmentations enhances the consistency among the scales.}\vspace{-3mm}
    \label{fig:main_pip}
\end{figure*}

We aim to design a generative adversarial network for image synthesis capable of: (a) full-frame or in-parts image generation, (b)  generation of arbitrary resolutions, and (c) consistency across different scales and parts.

\subsection{An image as a continuous space}
By viewing an image $I_z$ in a continuous coordinate space $\mathbb{R}^2$, image generation is seen as sampling image values at discrete locations within a finite rectangular area of this continuous space. We define the scale $s$ of the sampled image as the sampling period, and its resolution $r=(r_x, r_y) \in \mathbb{N}^2$ as the number of sampled points. Accordingly, the dimensions $(w, h)$ of the image in the continuous space are obtained as,
\begin{align}\label{eq:area}
(w, h) = (r_x * s_x, r_y * s_y) \in \mathbb{R}^2.
\end{align}
We also need a reference location for the rectangle in the continuous space to specify the rectangular region. We use the image's center coordinates $c=(c_x, c_y)$. Now, the tuple $a=(c, s, r)$ uniquely describes a sampled image $I_{z,a}$. Therefore, $I_{z,a}[i, j]$--the value for the pixel $(i, j)$ in $I_{z,a}$-- is obtained from the continuous image as:
\begin{align} \label{eq:pixel}
    I_{z,a}[i,j] = I_z(c_x + s_x i  - w/2, \nonumber\\
    c_y + s_y j  - h/2 ) \\
\end{align}

where $z$ is the semantic identifier of the image space. Each different scene/portrait/photograph has each unique $z$.

\subsection{Properties of Arbitrary-Scale Synthesis}

A convolution-based generator architecture needs specific characteristics to enable arbitrary-scale synthesis in a spatially equivariant and scale-consistent manner. This section will formulate these properties concerning the input positional encodings used as guidance.

\textbf{Position-guided generation}.
The generator needs to offer the ability to designate \emph{where} in the image space ($c$) and at which \emph{resolution} ($r$) and \emph{scale} ($s$) it should generate the image.
We give this information to the network via positional encodings $p_\text{enc}(a) = p_\text{enc}(c,r,s)$. 
Similar to the definition of $I_z$, each element of $p_\text{enc}$ designates a single location. The sampling period of the locations defines the scale, their number and alignment, the resolution.
$p_\text{enc}$ are different from the latent code $z$, which can be thought of as a description of the scene that produces another image space. 
Our generator network $G$ maps the latent code $z$ and a positional encoding as to an image space:
\begin{equation} \label{eq:1}
I_{z, a} = G(z, p_\text{enc}(a)), \text{where } a=(c,r,s)
\end{equation}

\textbf{Spatial equivariance} can be formulated as follows: A shift in reference location $c \xrightarrow{}  c'$ of the positional encodings should result in a similar shift in the image space,
\begin{equation} \label{eq:2}
I_{z, a'} = G(z, p_\text{enc}(a')) \text{where } a' = (c',r,s)
\end{equation} 

We can similarly define \textbf{scale consistency} as equivariance to the scale transformation  $s \xrightarrow{} s'$. 
\begin{equation} \label{eq:3}
I_{z, a"} = G(z, p_\text{enc}(a")) \text{where } a" = (c,r,s')
\end{equation} 

\subsection{Designing a scale- and translation-equivariant Generator}

We base our generator network $G$ on the commonly used StyleGANv2. First, we discuss the modifications needed to achieve the spatial and scale equivariance.
The generator's architecture  is mainly composed of a learned constant input, a modulated \xtimes 3 convolution layer, and $L$ blocks, each containing one upsampling layer and two modulated \xtimes{3} convolution layers.
The convolution layers use zero padding, which keeps the resolution of the input-output feature maps unchanged. 
The only size-changing operations in the generator are the up-sampling layers.
Let the input size be \xtimes{n_{in}}. The output resolution is given by:
\begin{equation} \label{eq:4}
 r_L = n_{in} * 2 ^ {L}
\end{equation} 
This means that the size of the output of a convolutional generator is directly proportional to the size of its input and can only have values with a $ 2 ^{L} $ increment.
To synthesize an image where the full-frame resolution is between two consecutive values, ~\eg $L_1, L_2$, $G$ needs to handle partial synthesis. 
The end result is either trimmed down $r_{L_2}$ or a stitched up version of smaller outputs. 

Both scale and translation equivariance are critical towards our objective.
As a generator architecture is a multi-step process, a natural way to impose equations \eqref{eq:2} and \eqref{eq:3} is for them to hold at each intermediate step.
Convolutional layers are, by design, translation equivariant. Thus, we address this property in  the rest of the components: \textit{padding}, \textit{upsampling} and \textit{positional input}.

\textbf{Removing the padding}. 
Zero padding breaks the translation equivariance of the network \cite{mspie,lin2021infinity}.
Removing it strips the network of its positional anchor.
Instead, positional encodings guide the generation of the image ~\cite{mspie,Choi2021ICCV}.
However, without padding, the \xtimes 3 convolution leads to a \emph{shrink-ed} output feature map compared to its input.

\textbf{Pitfalls when upsampling}.
One approach~\cite{mspie} to counter this shrinkage effect of zero padding,  
is to change the upsampling operation to a factor larger than two. 
This provides an excess of pixels in the feature maps that is subsequently consumed by the convolutions.
Specifically, as two convolution blocks are applied after the upsampling, the feature maps are resized from $n_{in}$ to $2*n_{in} + 4$.
However, this approach transforms the space unevenly when applied to different scales.
An input of \xtimes 4 will be rescaled with a factor of 3 while an input of \xtimes 8 with a factor of 2.5.

Similarly, an uneven transformation of space happens when resizing is done with aligned corners, both as part of the upsampling operation and the design of input encodings.

\textbf{Fixed positional corners}.
Xi~\ETAL{mspie} argue that using fixed values for the edge positional encodings, same for every scale, provides spatial anchors across the image space. 
While this is useful for generating images of specific set of scales, it impedes our arbitrary-scale and partial synthesis goal.
For translation equivariance, it is crucial for the encodings to point at the center of the pixel and not at the corners.
This way, two independently generated patches will be characterized by equally spaced positions. 
In multiscale synthesis, aligned-corners alter the sampling period between different scales,
where $d_{n \times n}  = (w/(n-1), h/(n-1))$.

Alternatively, sampling all the positional encodings as the central location of the patch they produce gives a period of  
$d_{n \times n}  = (w/n, h/n)$ and thus $2 d_{2n \times 2n}  = d_{n \times n}$.

\begin{figure}[t]
    \centering%
    \resizebox{0.80\linewidth}{!}{
 \includegraphics[width=\textwidth]{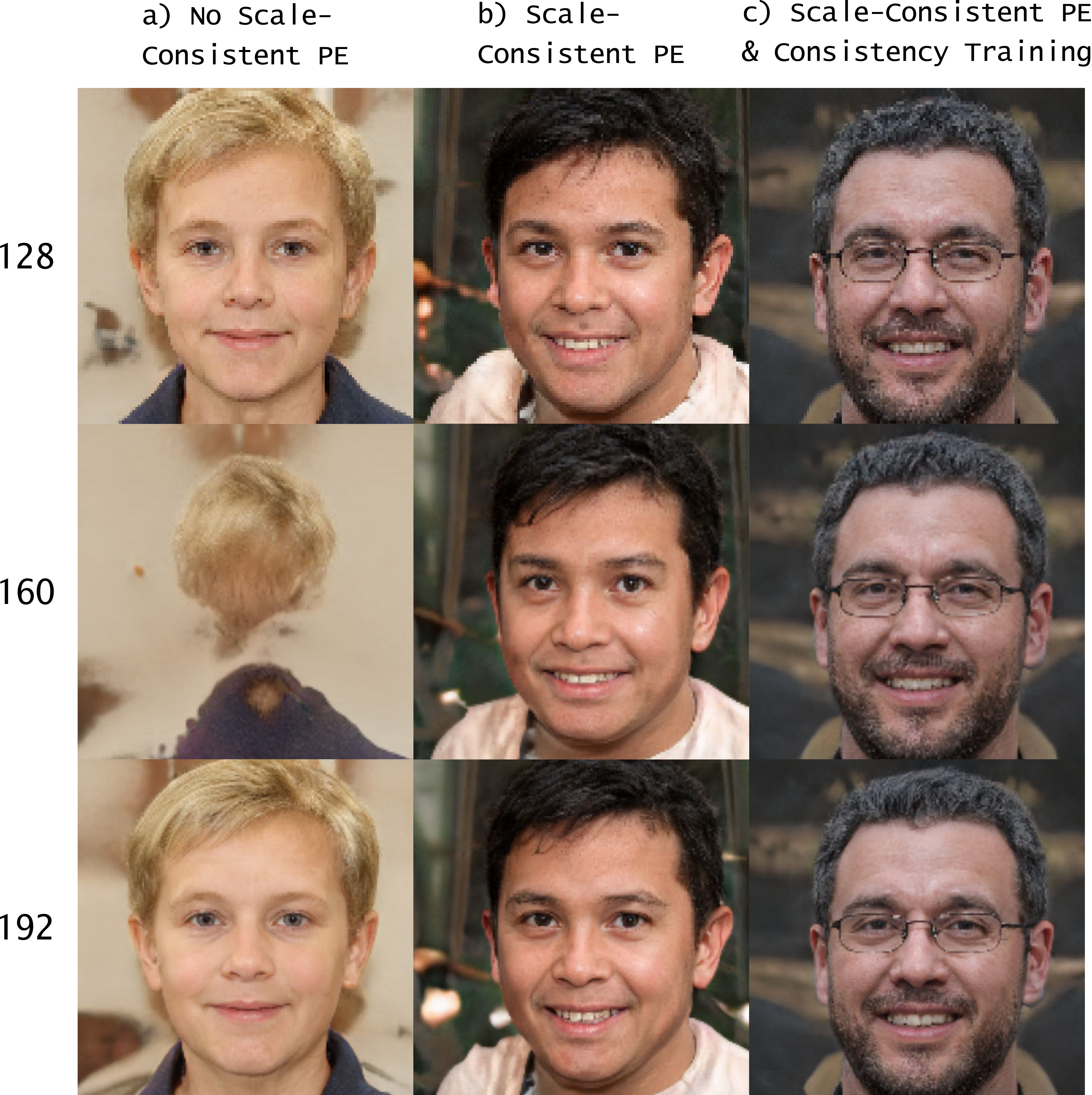} 
    }
 \caption{Results for different positional encodings and upsampling techniques. The generator was trained with two output resolutions: 128 and 192. Our positional encodings enable the generator to generate in a scale between the ones it was trained for, but it does not produce consistent results. Adding the scale-consistent objective and partial synthesis training alleviates this problem.}\vspace{-3mm}
     \label{fig:scale_comp}
 \end{figure}

This inter-scale inconsistency of the positional grounding between layers pushes the network to overfit to the scales it is trained to generate. Therefore, the generator is unable to synthesize in a scale in between. We can observe this effect in the first row of \fref{fig:scale_comp}.

\textbf{Scale consistent positional encodings}. 
We address the aforementioned issues in our design of the positional encodings.
A grid coordinate system is used as a natural and straightforward way to define them. 

As we want the positional encodings to describe the same area as the sampled image $I_{z,a}$, described by $a=(c,s,r)$ and the input resolution is $n \times n$, we find the sampling period to be $s_{n \times n} = (w/n, h/n)$ as per \eqref{eq:area}. 

To counter the shrinkage effect we utilize feature unfolding~\cite{chen2021learning, lin2021infinity}. 
However, for multiscale synthesis, the unfolding should be used as auxiliary padding and not taken into consideration when designing the encodings' sampling period to maintain $2 d_{2n \times 2n}  = d_{n \times n}$. 
Therefore, we extrapolate the positional encodings by the constant $n_\text{pad}$ on each side. We define the positional encodings as,
    \begin{align}
    \label{eq:7}
        p_\text{enc}(a)[i,j] = (c_x + s_x(i + 0.5)  - w/2,\nonumber\\
         c_y +s_y (j + 0.5) - h/2 ), \nonumber \\
        \forall i,j \in [-n_\text{pad}, n+n_\text{pad}) \cap \mathbb{Z}   
    \end{align}
Note that $n_\text{pad}$ does not affect the scale $s$. Using $p_\text{enc}(a)$ as the input to our StyleGAN2-based architecture the resolution of the intermediate feature maps is:
\begin{align}
\label{eq:5}
 n_{out}^0 &= n + 2 n_\text{pad} - 2 && \text{\small For the first convolution} \nonumber \\
 n_{out}^l &= n_{out}^{l-1} * 2 - 4 && \text{\small For each upsampling block}
\end{align} 

By setting $n_\text{pad} = 3$ we get:
\begin{align}
\label{eq:6}
 n_{out}^l = n_{in} * 2 ^ {l} + 4
\end{align} 
These extra 4 pixels at the margins of each intermediate feature map are there \textit{regardless} of the input size. They play the auxiliary role of keeping equation~\eqref{eq:6} consistent among layers. We remove them at the end of the network, and thus our output resolution is described by the same formula as its zero-padded counterpart (Equation \eqref{eq:4}).

Feature unfolding designates an image area larger than the one we want to generate. 
The upsampling is doubling the scale without changing the area.
The convolutions consume the excess area, but the area described by the initial positional encodings does not change between layers. 

A shift in positional encodings translates to a shift of the image. Additionally, changing the spacing between them without increasing their number will change the size of the area they describe and let us generate a continuum of scales.

\subsection{Training for scale}

While the design choices described in the previous subsection permit the generation of arbitrary-sized images, they do not guarantee consistency among images generated from the same latent code but at different scales.  To achieve this, we propose a scale consistency objective.

\textbf{Training pipeline.} In order to train for a multiscale objective, we teach the generator to synthesize images of different scales. 
For each batch, we randomly choose the output resolutions $r_\text{small}$ and  $r_\text{large} = 1.5 * r_\text{small}$ from a predefined set, in accordance with Equation \eqref{eq:6}.

Assuming a generator with 6 upsampling blocks,
we pick $r_\text{small} = 256$ and $r_\text{large} = 384$.  This gives us $n_\text{small} = 4$ and $n_\text{large} = 6$.
Then, we randomly choose the scale $s$ of the image that will be generated and its location $(c_x, c_y)$. Lastly, we sample the latent code $z$.
Thus, we get,
\begin{align}
\label{eq:8}
        I_{z,a_\text{small}} &= G(z, p_\text{enc}(a_\text{small})) \\
        I_{z,a_\text{large}} &= G(z, p_\text{enc}(a_\text{large}))
\end{align}
Similarly, we crop and resize the real images per $(c_x, c_y)$.

\textbf{Scale consistency.}
The classic adversarial training only pushes images to look realistic at each scale.
We need to define an objective that will teach the generator to match the outputs. 
A straightforward approach is to impose a distance metric, such as L1 loss, between images generated at different scales and subsequently resized to match.
However, this can give the network conflicting incentives.
The L1 loss drives the different images to match without any regard to their perceptual quality; two uniformly black images would achieve the perfect L1 loss.

We propose a scale consistency approach that strives simultaneously to generate similar images at different scales and images that look realistic.
To achieve this, we use augmentation techniques during the training of the discriminator without changing its loss function. 

We deploy two types of augmentations before feeding
$I_{z,a_\text{small}}$ and $I_{z,a_\text{large}}$ to the discriminator.
First,
we use \textit{CutMix}\cite{cutmix} to crop a region at one scale and substitute it with a resized crop of the same region of the image generated at the other scale.
Then, we use ChannelMix to randomly substitute some of the RGB channels of the image at each scale with ones from its counterpart, after it is resized to match.

The discriminator is trying to measure the realness of the mixed images.
In the process, the generator learns to associate the identity of images it synthesizes with the style code
and their position and scale with the input positional map. 
The whole pipeline of our method is shown in \fref{fig:main_pip}.

\textbf{Global consistency for partial generation by Multiscale Training}.
Combining partial and multiscale training naturally counters a common partial synthesis problem: global consistency. The generator can create a consistent large resolution full-frame image at inference, without explicitly trained for it. The network learns the global structure by being taught to generate small-resolution full-frame images, and detailed textures of high-resolution patches.

\textbf{Handling Injected Noise during inference}.
In Equation \eqref{eq:1} we described a simplified formulation of the generator that omitted the injected noise at the end of each convolution. 
We strive for consistency among different scales of images produced with the same latent code, but randomly sampling the injected noise works against this objective.

Imposing scale-consistent positional encodings enables a practical feature. We know the positional grounding of every pixel of every intermediate feature map. This lets us have a position-aware interpolation of the noise to match corresponding pixels between scales. 

Similarly, the same technique can be used towards translation equivariant synthesis.
We shift the intermediate noise according to the positional encodings' shift and sample only the portion of the images outside the generational frame.

%% file: source/experiments.tex
\begin{figure}
    \centering%
    \resizebox*{1.0\columnwidth}{!}{
    \includegraphics[width=1\textwidth,trim= 0 0 0 0]{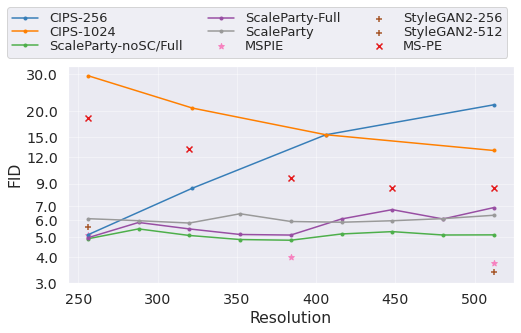}
    }
    \caption{FID scores for entire face generation for a continuum of scales for the FFHQ dataset. The continuous lines indicate the methods that can generate in arbitrary scales. ScaleParty performs competitively to single-scale models.}\vspace{-1mm}
    \label{fig:fid-res}
\end{figure}
\begin{table}[]
    \centering
    \resizebox*{0.8\columnwidth}{!}{
    \begin{tabular}{c|cccc}
     & \multicolumn{4}{c}{Self-SSIM(5k)}  \\
     Method: & 320 & 384 & 448 & 512 \\
        \hline
MSPIE\cite{mspie} & 0.1194&0.5929&0.3316&0.5785 \\
MS-PE\cite{Choi2021ICCV} & 0.9128&0.8687&0.8367&0.8112 \\
CIPS-256\cite{anokhin2020cips} & \textbf{0.9991}&\textbf{0.9987}&\textbf{0.9985}&\textbf{0.9981} \\
ScaleParty-noSC/Full & 0.7154&0.6975&0.6489&0.6511\\
ScaleParty-Full & 0.8637&0.8942&0.8266&0.8114 \\
ScaleParty & 0.8802&0.8779&0.8568&0.8454 \\

    \end{tabular}}
    \caption{Self-SSIM between 5k FFHQ generated images of different scales, resized and compared at resolution  \xtimes{256}.}\vspace{-3mm}
    \label{tab:selfssim}    
\end{table}

\section{Experimental Results}
\label{sec:experiments}

\subsection{Implementation}
We base our implementation\footnote{Code: \url{https://github.com/vglsd/ScaleParty}} on MS-PIE\cite{mspie} using the \textit{mmgeneration} framework\cite{2021mmgeneration} built upon PyTorch\cite{pytorch}.
For all upsampling operations, we use bilinear interpolation without corner alignment.
In order to match the feature maps of the network's RGB branch, we remove the feature maps' marginal pixels  after upsampling.
Our model is trained with the non-saturating logistic loss, with R1 gradient penalty~\cite{pmlr-v80-mescheder18a} for the discriminator and path regularization for the generator~\cite{Karras2019stylegan2}.
We used the StyleGAN2 discriminator~\cite{Karras2019stylegan2} together with an adaptive average pooling layer before the last linear layer~\cite{mspie,he2014spatialpyramid}. For all our experiments we set $h=w=2$ for the encodings calculation. 
\input{source/experiments_source/evaluation.tex}
\input{source/experiments_source/results.tex}
\subsection{Applications}
\input{source/experiments_source/transformations.tex}
\input{source/experiments_source/projections.tex}

%% file: source/experiments_source/evaluation.tex
\subsection{Evaluation}

\textbf{Datasets.}
We evaluate using three different datasets:
\begin{itemize}\vspace{-2mm}
    \item The \textit{Flick-Faces-HQ} (FFHQ)~\cite{Karras_2019_CVPR} is composed of 70,000 images of diverse human faces. This dataset is characterized by a strong positional prior as the images are cropped and aligned from photographs with a larger context, based on facial landmarks. The original size of the pictures is \xtimes{1024}. We train on FFHQ by cropping and then downsampling the images. \vspace{-3mm}
    \item The \textit{LSUN} dataset ~\cite{yu15lsun} consists of  images that are resized, so their smaller side is 256 pixels. We test our method in two subcategories of the dataset: 
    the \textit{LSUN Bedroom}, which consists of  3 million bedroom images and 
    the \textit{LSUN Church}, which has 126 thousand diverse outdoor photographs of churches. While each dataset depicts a similar layout of bedroom and outdoor churches scene, the positional priors of the images are not as strong as in FFHQ. To further reduce their strength we randomly crop square patches of the images while training, without altering the aspect ratio.
\end{itemize}

\textbf{Metrics.}
We rely on commonly used metrics to measure two aspects of multiscale-generation.
Frechet Inception Distance~\cite{NIPS2017_7240} assesses the perceptual quality at each scale. It is shown to align with human subjects' perceptual judgement of an image. Improved Precision and Recall~\cite{precisionrecall} is used to gauge the plausibility of the synthesized images and how well these images cover the range of the distribution of the real images, respectively.
To assess the consistency among images generated at different scales, we  deploy the SSIM metric. We call it SelfSSIM. 

Note that consistency on its own should not be the goal: two equally bad syntheses can have high fidelity among them. SelfSSIM is used with FID to assess whether the generated images are perceptually good and consistent.

%% file: source/experiments_source/results.tex
\subsection{Quantitative Results}
\begin{table}[]
    \centering
    \resizebox*{\columnwidth}{!}{
    \begin{tabular}{c|c|ccc|ccc}
    Method & Res  & FID & Prec & Rec & \multicolumn{3}{c}{SelfSSIM (5k)}  \\
        \hline
    Dataset: & \multicolumn{5}{c}{LSUN Church}  \\
        \hline
        
        MSPIE\cite{mspie} & 128 & \textbf{6.67} &\textbf{ 71.95} & \textbf{44.59} & 1.00 & 0.32 & 0.43 \\
        & 160 & 10.76 & 66.21 & 36.95 & 0.31 & 1.0 & 0.40 \\
        & 192 &\textbf{ 6.02} &\textbf{ 66.70} & \textbf{46.16} & 0.39 & 0.38 & 1.00 \\
        ScaleParty-noSC/Full & 128 & 7.62 & 70.21 & 39.84 & 1.00 & 0.58 & 0.49 \\
        & 160 &\textbf{ 7.47} & \textbf{72.23} & \textbf{39.44} & 0.55 & 1.00 & 0.67 \\
        & 192 & 7.40 & 67.83 & 39.93 & 0.44 & 0.64 & 1.00 \\
        Scaleparty & 128 & 9.08 & 70.52 & 32.10 & 1.00 & \textbf{0.95} & \textbf{0.93} \\
        & 160 & 7.96 & 70.87 & 32.07 & \textbf{0.94} & 1.00 & \textbf{0.95} \\
        & 192 & 7.52 & 68.14 & 33.33 & \textbf{0.90} & \textbf{0.94} & 1.00 \\

       \hline
    Dataset: & \multicolumn{5}{c}{LSUN Bedroom}  \\
        \hline
        MSPIE\cite{mspie} & 128 & 11.39 & \textbf{66.45} &\textbf{ 26.97} & 1.00 & 0.10 & 0.10 \\
        & 160 & 16.45 & 63.84 & 23.09 & 0.10 & 1.00 & 0.12 \\
        & 192 & 12.65 & 58.10 & 25.93 & 0.10 & 0.12 & 1.00 \\
        ScaleParty-noSC/Full & 128 & 11.45 & 63.26 & 25.42 & 1.00 & 0.67 & 0.55 \\
        & 160 & 10.80 & \textbf{64.48} & \textbf{25.77} & 0.64 & 1.00 & 0.75 \\
        & 192 & 11.56 & 60.87 & \textbf{26.64} & 0.50 & 0.73 & 1.00 \\
        ScaleParty & 128 &\textbf{ 10.15} & 62.50 & 20.63 & 1.00 & \textbf{0.94} & \textbf{0.92} \\
        & 160 & \textbf{9.85} & 64.14 & 22.02 & \textbf{0.92} & 1.00 & \textbf{0.95} \\
        & 192 & \textbf{9.92} & \textbf{64.77} & 21.10 & \textbf{0.89} & \textbf{0.94} & 1.00 \\

    \end{tabular}}
    \caption{Evaluation Metrics on LSUN Church and Bedroom datasets~\cite{yu15lsun}. The datasets do not exhibit strong positional prior, which increases the performance gain of our approach.}\vspace{-3mm}
    \label{tab:lsun}    
\end{table}

\begin{figure}
   \centering 
   \noindent\includegraphics[width=0.45\textwidth, trim=0 20 0 50]{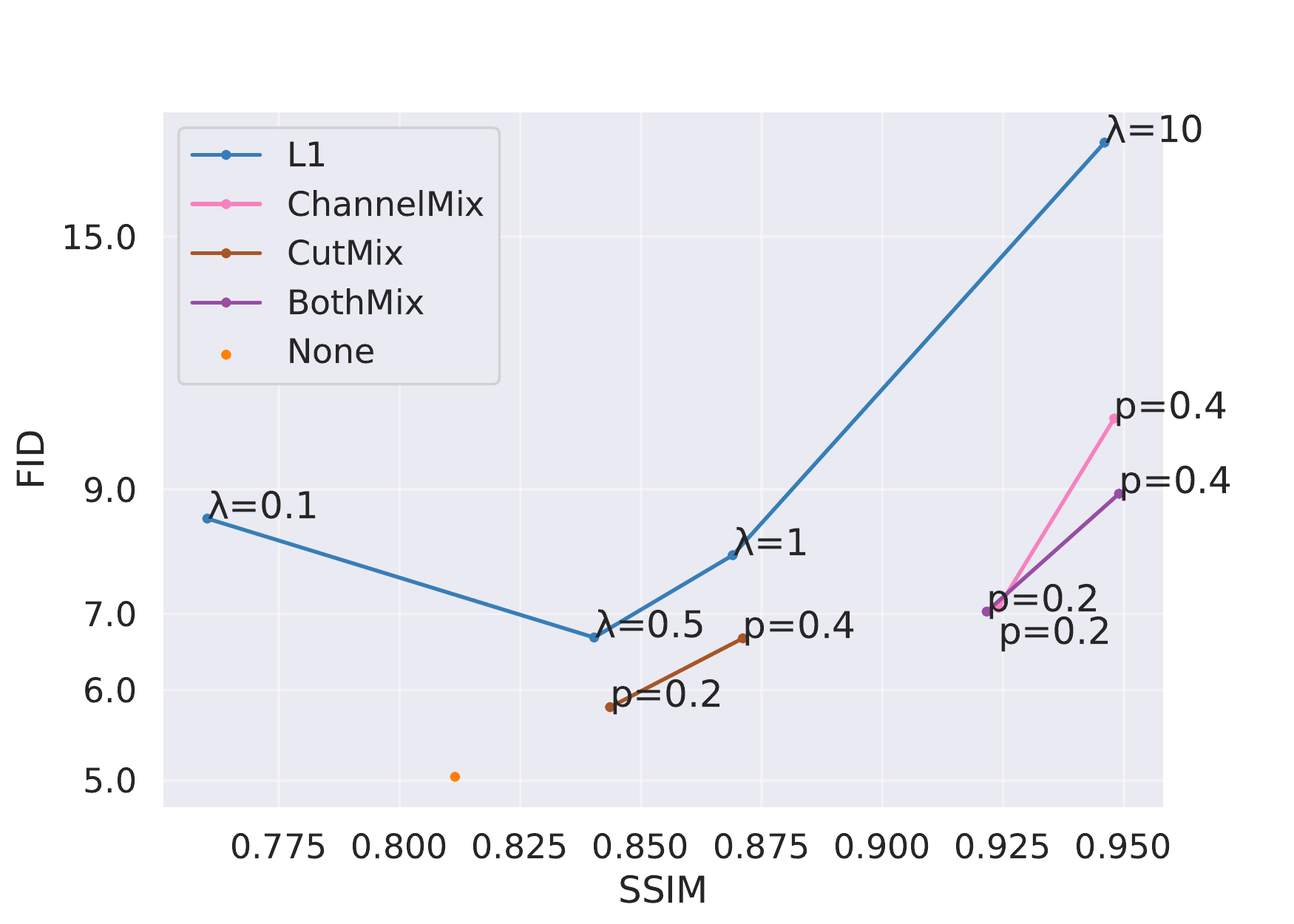} 
\caption{Trade-off between FID and self-SSIM on FFHQ. Enforcing stricter scale consistency leads to a drop in perceptual quality. $\lambda$ indicates the weight applied to $L1$ loss, while $p$ the probability to apply our inter-scale augmentations to a training batch.} \vspace{-3mm}
\label{fig:tradeoff}
\end{figure}

\textbf{Comparison with state-of-the-art models on FFHQ}.
We use the FFHQ dataset to conduct a comparative analysis with state-of-the-art methods in multiscale generation. We test against methods designed for multiscale synthesis: MSPIE~\cite{mspie} and MS-PE~\cite{Choi2021ICCV}. From the INR-based methods, we compare against CIPS~\cite{anokhin2020cips} as it reports better FID than INR-GAN~\cite{inr_gan} and their implementation readily handles synthesis at arbitrary scales. We report the results of two models: one trained for \xtimes{256} and one trained for \xtimes{1024} images. Lastly, we include the instances of the single-scale StyleGAN2\cite{Karras2019stylegan2} model as a benchmark.

In \fref{fig:fid-res} and \tref{tab:selfssim} we are reporting the FID scores and the SelfSSIM scores respectively. To calculate both metrics we did not use the truncation trick.

\textit{ScaleParty vs. other methods}. Only StyleGAN \& MSPIE consistently yield better FID scores than our approach. However,  they overfit the set of scales they were trained for and  are incapable of good syntheses outside this set. CIPS has a competitive score for the single scale it was trained for, which rapidly deteriorates as we move away from that scale. CIPS has the best SelfSSIM. Note, CIPS is conditioned on a single location that does not contain any scale information. Generating in higher scales could emulate a naive upsampling method, which similarly would yield almost perfect SelfSSIM. Therefore, ScaleParty is the only method that can consistently achieve low FID scores while maintaining high inter-scale consistency.

\textbf{The effects of ScaleParty components on FFHQ}. We train and compare with two versions of our model, ablating on our proposed elements: (a) \textbf{ScaleParty-noSC/Full} is trained with our proposed scale-invariant positional encodings, but only with full-frame images of a discrete set of scales and no consistency objective.  (b) \textbf{ScaleParty-Full} is trained with full-frame images and an additional scale-consistency objective: in 20\% of the batches, we generate a multiscale pair of images. In contrast, our full model ScaleParty is trained with both the scale-consistency objective \textit{and} for partial generation. During training, the positional encodings (and real images respectively) are sampled with a scale $60-110\%$ of the full-frame.

We find that increasing the inter-scale consistency comes at a slight drop in perceptual quality. As seen in ~\fref{fig:fid-res}, ScaleParty-noSC/Full produces the best FID score compared to the configurations imposing scale consistency. Partial generation trains the generator for different scales. While ScaleParty-Full results in better SelfSSIM for the trained full image resolutions, we observe a drop in consistency for the scales the network was not trained with. 
However, upon visual inspection we notice an unnatural distortion in the faces generated without partial synthesis training, that is not reflected in the FID, as seen in Figure 4 of the supplementary material for both ScaleParty-noSC/Full and ScaleParty-Full. This distortion explains the lower SelfSSIM between the unseen and the trained-for scales. 

\begin{table}[]
    \centering
    \resizebox*{0.7\columnwidth}{!}{
    \begin{tabular}{c|cccc}
     & \multicolumn{4}{c}{Self-SSIM(5k)}  \\
     Method: & 279&307&341&384 \\
        \hline
Random & 0.8648&0.8546&0.8310&0.8501 \\
Constant & 0.8678&0.8558&0.8389&0.8479 \\
GridSample & \textbf{0.8960}&\textbf{0.8826}&\textbf{0.8603}&\textbf{0.8712} \\

    \end{tabular}}
    \caption{The effect of the sampling noise method on SelfSSIM-\xtimes{256}.
    Our proposed grid sampling based on the relative position of each pixel of each intermediate feature map, yields an improvement between 0.02 and 0.03 compared to naive approaches.
    }\vspace{-3mm}
    \label{tab:noisessim}
\end{table}

\textbf{The effects of ScaleParty components on LSUN Dataset}. In contrast to FFHQ, LSUN lacks strong positional priors. The difference is intensified due to the random cropping. For investigating this setting we train MSPIE as our baseline, as it also deploys a pad-free generator. Furthermore we train ScaleParty-noSC/Full along with our main configuration, ScaleParty, to illustrate the benefits of our scale-invariant design and our scale consistency objectives respectively.
In Table \ref{tab:lsun} we can see the results.

MSPIE and ScaleParty-noSC/Full are trained with \xtimes {128} and \xtimes {192} full-frame images, sampled with equal chance. The inconsistency between the positional encodings hinders MSPIE's association of the positional input to the output. Both noise injection and generation at different scales lead to a change of the location of the generated images, resulting in a poor SelfSSIM, even with good FID. In contrast, our positional encodings learn the association, enabling good synthesis at the unseen resolution of 160.

Compared to the positionally structured FFHQ where MSPIE and ScaleParty-noSC/Full achieve relatively high SelfSSIM, these configuration exhibit poor consistency in LSUN datasets. In contrast, our ScaleParty shows similarly high results, recording even higher increase compared to the face dataset.
We refer to the supplementary material for visual comparisons on both FFHQ and LSUN datasets.

\textbf{Ablation on scale consistency approaches.}
We investigate the effect scale consistency has on the perceptual quality of the generated images.
We start with our ScaleParty-noSC/Full model, where no such objective is imposed. We experiment applying L1 loss and combinations of our suggested inter-scale augmentations: CutMix~\cite{cutmix} and ChannelMix. For L1 loss we test for $\lambda$ values of 0.1, 0.5, 1.0 and 10. Then, we ablate on the augmentations by altering whether and how often they are applied per iteration.

In \fref{fig:tradeoff} the trade-off between SelfSSIM and FID is visualized. For brevity, we calculate the FID for \xtimes{128} resolution images and the SSIM between images of \xtimes{128} against downsized \xtimes{192} images.

The L1 experiments yield worse SelfSSIM for the same perceptual scores compared to our proposed augmentations approach. The difference is intensified for larger $\lambda$.
When consistency increases, perceptual quality decreases.
ChannelMix is pushing for global consistency compared to CutMix where the network needs to stitch the two images. The increase of the frequency of scale-consistency batches ($p=0.2$ vs $p=0.4$) increases both metrics.

\textbf{How to sample the injected noise.}
    While our network strives to produce consistent results across scales there is a form of randomness that we have not addressed until now: convolutional noise. We experiment with three different policies for sampling the noise across scales. 
         \textit{a) Random:} the noise is randomly sampled at each layer of each scale.
         \textit{b) Constant:} the noise is only sampled at the largest scale and reused for generating each smaller scale.
         \textit{c) GridSample:} the noise is only sampled at the largest scale. Then, we utilize the scale consistent positional encodings to interpolate the sampled values to smaller scales. 
         
 For  fair comparison, we run this experiment three times and report the average SelfSSIM. For each run 1000 style codes are shared among the different policies. Moreover, we sample a single noise map for both \textit{Constant} and \textit{GridSample}.
    The images were resized and compared at resolution  \xtimes{256}. The proposed grid sampling method outperforms the other approaches as shown in \tref{tab:noisessim}.
    

%% file: source/experiments_source/transformations.tex
\begin{figure}[t]
    \centering%
    \resizebox*{1.0\columnwidth}{!}{
    \includegraphics[width=1.\linewidth, trim=40 00 0 00]{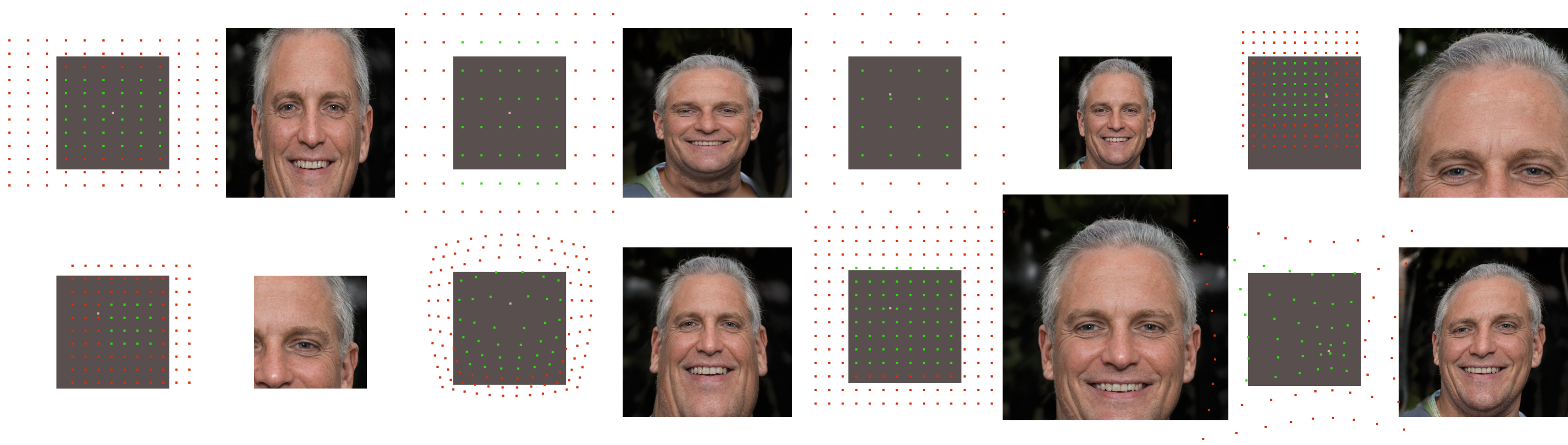}  
    }\vspace{-3mm}
    \caption{Transformations of the positional encodings result to the equivalent transformation of the output image.
    While the network was not trained for this, partial and multiscale training enables the generator to generalize to unseen input configurations.}\vspace{-3mm}
    \label{fig:transforms}

\end{figure}

\textbf{Geometric manipulation using positional encodings.}
Training for both multiscale and partial synthesis requires the convolutional generator to learn to interpret a great variety of positional encodings configurations. 
We present it with unseen configurations to test how well the generator learned to translate the positional input.
We show qualitative results of applying transformation on the input positional encodings. 
 In \fref{fig:transforms} and \fref{fig:intro}c  we can observe : (a) transformation of the \textbf{aspect ratio},
(b) \textbf{warping},
(c) \textbf{unseen resolutions} 
and (d) \textbf{extrapolation.}

%% file: source/experiments_source/projections.tex
\textbf{Projection of real images.}
We investigate the ability of our network to represent real images within its latent space.
Following Abdal~\ETAL{embedstylegan} we deploy optimization of the style vectors modulating each layer of our network ($W^{+} space$).
We aim to minimize the perceptual~\cite{zhang2018perceptual} and $L_2$ loss between the real and generated images while keeping the weights of the generator frozen.

\begin{figure}[t]
    \centering%
    \includegraphics[width=1.000\linewidth, trim=0 0 20 120, clip]{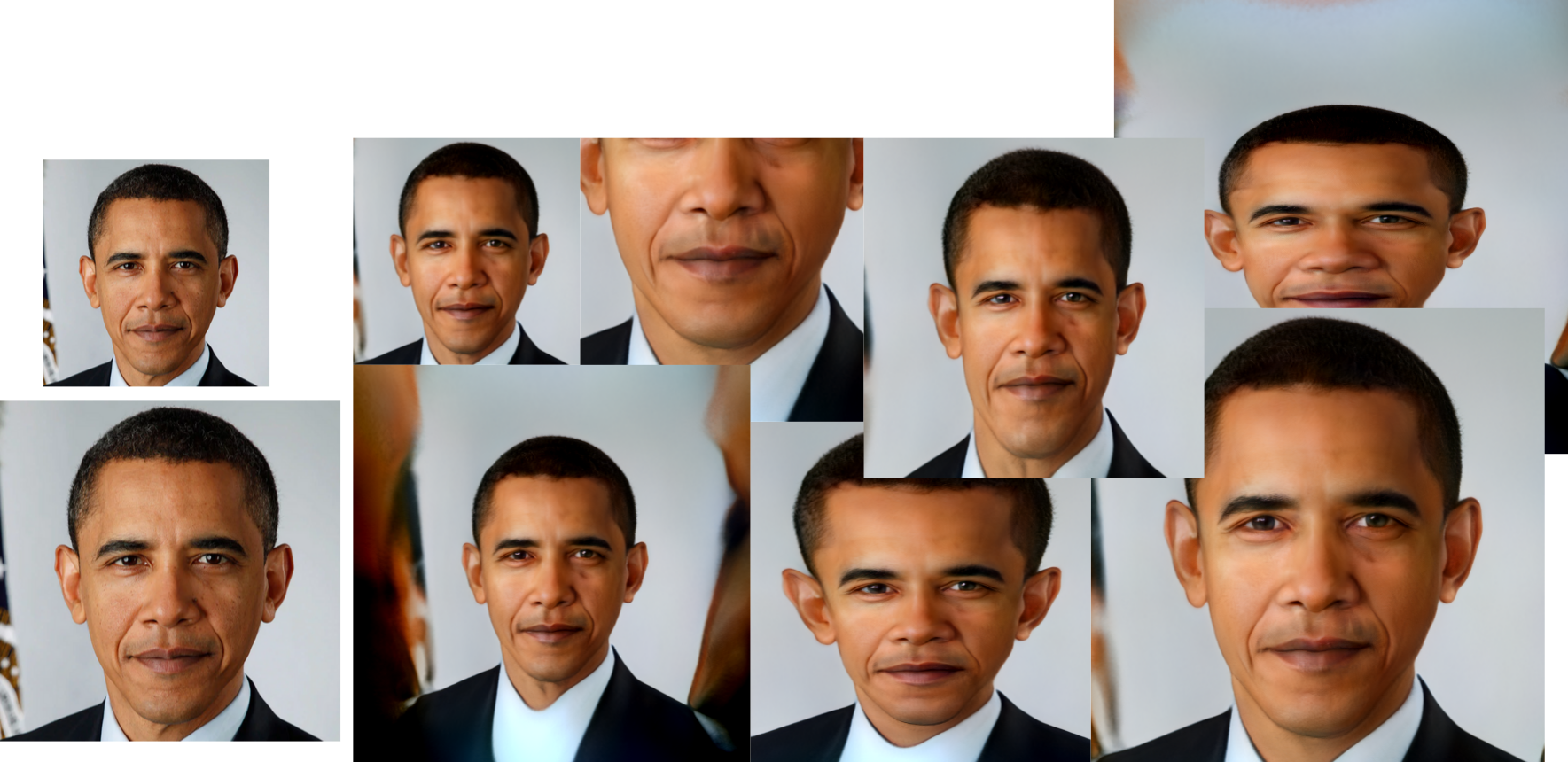}\vspace{-3mm}
    \caption{A single latent code is optimized to match the two real images on the left. We apply geometric transformations to positional encodings to generate various images conditioned on the inferred latent code. Note that only the input positions are transformed, therefore we circumvent the pixelation effect that these transformations would cause if applied on the image space. }\vspace{-3mm}
    \label{fig:realtransforms}
\end{figure}

We find that optimizing the latent code for a single-scale image leads to scale overfitting. However, by optimizing the same latent code for two scales simultaneously we are able to also generate in all scales in between. In \fref{fig:realtransforms} we use the repertoire of transformations described in the previous subsection to geometrically manipulate a real images.

%% file: source/conclusion.tex
\section{Conclusion}
\label{sec:conclusion}

We present ScaleParty, a novel method for Arbitrary-Scale Image Synthesis utilizing a single generative adversarial network trained with positional guidance.
We show that our scale-consistent positional encodings permit a pad-free generator to produce perceptually good results across a continuum of scales.
Furthermore, we introduce a scale-consistency objective by applying inter-scale augmentations before presenting the synthesized image to the discriminator network.
Incorporating partial generation training in our pipeline further improves consistency. 
The combination of multi-scale and partial synthesis training teaches the generator a dense representation of positional encodings.
During inference, this can be leveraged to create geometrically manipulated images by applying transformations such as warping or stretching to positional encodings.

%% file: source/societal_impact.tex
\section{Societal Impact}
The growth of deepfakes appearing online is a cause for serious concern in a multitude of domains including: politics, non consensual usage of data and the general feeling of losing faith in digital information. GANs are the main technological advancement that enabled the rise of this content. The presented work does not directly lend itself to creation of fake material, in the sense of replacing faces or creating facial expressions based on audio stream. Indirectly though, our method can be used to geometrically manipulate images and in this sense provide malevolent users an additional tool. Efforts in both the US (S. Rept. 116-289 - IDENTIFYING OUTPUTS OF GENERATIVE ADVERSARIAL NETWORKS ACT) and the EU(2021/0106(COD) Artificial Intelligence Act ) are aiming to legislate the creation of deepfakes, while private companies try to detect and ban the spreading of deepfake material on the internet. 
Our method gravitates towards white colored faces in the center of the latent space due to the imbalance on the used data set. There is a clear need to create diverse data sets, where people are represented equally independent of their ability to access technological resources. This will enable research to be used in a more wide spectrum of applications across the globe. 
In terms of the ever increasing computational costs of training deep neural networks, our presented method overcomes the need for creating independent models at each resolution. It can be therefore be used to reduce the required energy by replacing multiple single resolution models with a single scale consistent one.

%% file: source/limitations.tex
\section{Limitations}
We use artificially multi-scale datasets to train ScaleParty.
We downsample the images to acquire different scales.
Parmar~\etal~\cite{parmar2021cleanfid} argue that different resizing libraries and methods can have drastic effects on the quality of the resized images. This is an aspect we have not investigated.  
In equation (5) of the main paper we assume that transitive closure applies to resizing, e.g. resizing from \xtimes{512} to \xtimes{256} is equivalent to resizing to \xtimes{384} as an intermediate step. While this assumption is not true, it still helps us with our scale-consistency objective.
Nevertheless, an analysis on a  naturally multi-scale dataset would greatly benefit the conclusions of this work.
